\documentclass{article}

\usepackage{PRIMEarxiv}

\usepackage[utf8]{inputenc} 
\usepackage[T1]{fontenc}    
\usepackage{hyperref}       
\usepackage{url}            
\usepackage{booktabs}       
\usepackage{amsfonts}       
\usepackage{nicefrac}       
\usepackage{microtype}      
\usepackage{lipsum}
\usepackage{fancyhdr}       
\usepackage{graphicx}       
\graphicspath{{media/}}     

\usepackage{verbatim}
\usepackage{algorithm,algpseudocode}
\usepackage{todonotes}
\usepackage{amsmath}

\title{Resilience from Diversity: Population-based approach to harden models against adversarial attacks}

\author{
  Jasser Jasser \\
  Department of Computer Science \\
  University of Central Florida \\
  Orlando, FL, USA\\
  \texttt{Jasser.Jasser@ucf.edu} \\
   \And
  Ivan Garibay \\
  Department of Industrial Engineering and Management Systems \\
  University of Central Florida \\
  Orlando, FL, USA\\
  \texttt{igaribay@ucf.edu} \\
}

\begin{document}
\maketitle

\begin{abstract}
Traditional deep learning networks (DNN) exhibit intriguing vulnerabilities that allow an attacker to force them to fail at their task. Notorious attacks such as the Fast Gradient Sign Method (FGSM) and the more powerful Projected Gradient Descent (PGD) generate adversarial samples by adding a magnitude of perturbation $\epsilon$ to the input's computed gradient, resulting in a deterioration of the effectiveness of the model's classification. This work introduces a model that is resilient to adversarial attacks. Our model leverages an established mechanism of defense which utilizes randomness and a population of DNNs. More precisely, our model consists of a population of $n$ diverse submodels, each one of them trained to individually obtain a high accuracy for the task at hand, while forced to maintain meaningful differences in their weights. Each time our model receives a classification query, it selects a submodel from its population at random to answer the query. To counter the attack transferability, diversity is  introduced and maintained in the population of submodels. Thus introducing the concept of counter linking weights. A Counter-Linked Model (CLM) consists of a population of DNNs of the same architecture where a periodic random similarity examination is conducted during the simultaneous training to guarantee diversity while maintaining accuracy. Though the randomization technique proved to be resilient against adversarial attacks, we show that by retraining the DNNs ensemble or training them from the start with counter linking would enhance the robustness by around 20\% when tested on the MNIST dataset and at least 15\% when tested on the CIFAR-10 dataset. When CLM is coupled with adversarial training, this defense mechanism achieves state-of-the-art robustness.

\end{abstract}


Deep learning models have achieved great success in computer vision applications and have become the preferred tool for solving a variety of problems. 
In academia, deep learning is heavily used in image generation, classification, and segmentation. 
Across industries, deep learning excels in self driving cars, healthcare, fraud detection, and many more application domains. However, recent literature \cite{szegedy2013intriguing} has shown that these models have an intriguing property that would render them vulnerable to adversarial attacks. 
There is a plethora of attacks that can target and harm an already trained deep learning model and force it to misclassify images that looks normal to the human eye.  
This work focuses on two popular attacks that are a standard in research, the FGSM \cite{goodfellow2014explaining} and PGD \cite{madry2017towards} attacks. 
In the FGSM attack, a one-step update along the sign of the gradient of the loss is executed to achieve the most increase in the loss, as shown in equation(\ref{eq:fgsm}).
The perturbation magnitude $\epsilon$ is added to the sign of the gradient of the loss function $J$ with regards to $x$. 
$\theta$ is the model's parameters.

\begin{equation}
  x' = x + \epsilon \cdot sign(\nabla_x J(\theta,x,y))
  \label{eq:fgsm}
\end{equation}

while in the PGD attack, maximizing the adversarial loss is calculated using smaller steps over several iterations and from a random starting point, shown in equation(\ref{eq:pgd}).
$x'$ is the perturbed sample at time $t$, $\prod_{x+S}$ is the perturbation projection towards $S$ centered around $x$, and $\alpha$ is the step size.

\begin{equation}
  x'_{t+1} = \prod_{x+S}{(x'_t + \alpha \cdot sign(\nabla_x J(\theta,x'_t,y)))}
  \label{eq:pgd}
\end{equation}

Currently, the most prominent and effective approach \cite{madry2017towards} against such adversary is by training the models with adversarial samples over a min-max objective as shown in equation(\ref{eq:def}) where the function $J$ is the adversarial loss, and the function $D$ is the distance between the benign example and the randomly selected starting point.

\begin{equation}
  \min_\theta \max_{D(x,x')<\eta} J(\theta,x',y)
  \label{eq:def}
\end{equation}

This work introduces a unique approach to defending against adversarial attacks.
Our approach relies on a basic principle from previous work \cite{zhou2018breaking} that "the adversary does not have unlimited attack budget."
Therefore creating randomness in a population of DNN submodels "eliminates the possibility of full knowledge acquisition."
It also disrupts the transferability of adversarial samples.
Our system consist of a population of models.
Each model is trained independently for accuracy while using diversity-preserving mechanism that results in submodels with comparable accuracy but different weights.
The system is flexible and can produce robustness without adversarial training, yet it can be trained with adversarial samples.
We consider defense strategies with and without adversarial training and show that our approach improves both.

Adversarial training-based defenses improve the robustness of deep learning models by training them with adversarial samples. Current research \cite{goodfellow2014explaining, madry2017towards} shows its effectiveness on all attack environments whether black-box, gray-box, or white-box. In FGSM adversarial training, Goodfellow et al. \cite{goodfellow2014explaining} train their model with benign and FGSM generated adversarial samples to robustify it. PGD adversarial training \cite{madry2017towards} does similar job where PGD perturbed samples are used to train the model adversarially. In this work, we adapt the PGD adversarial training to create a model consisting of PGD adversarially trained diverse submodels where diversity is enforced by counter linking (CLM-Adv).

Heuristic defenses that are based on randomization and denoising achieved a significant robustness against black-box and gray-box environments. In randomization, there are defenses that does random transformation to the input \cite{xie2017mitigating, guo2017countering}, or adding a random noise before every convolutional layer during the training and testing process \cite{liu2018towards, lecuyer2019certified, li2018certified}, or feature pruning \cite{dhillon2018stochastic} where a subset of the layers' activations with lower magnitudes are pruned and those with higher magnitudes are scaled up to compensate. In denoising, there are two types: input denoising, and feature denoising. In input denoising \cite{xu2017feature, xu2017feature2}, the defense tries to remove the adversarial perturbation in a straightforward fashion. While in feature denoising \cite{liao2018defense, athalye2018robustness}, the features affected by the attack's perturbation are polished by an image denoiser based on the high-level representation to suppress the influence of adversarial perturbations.

Provable defenses which are theoretically proven to counter adversarial attacks with a certain accuracy, depending on the attack class. Semidefinite programming-based defenses \cite{raghunathan2018certified, raghunathan2018semidefinite} outputs an optimizable certificate that encourages robustness against all attacks.  The defense proposed in \cite{wong2018provable}, considers a convex outer approximation that covers all the perturbations that can possibly be generated from that space and minimizes the worst-case loss over this region through linear programming. The work of \cite{guo2018sparse, hein2017formal, weng2018evaluating, xiao2018training} investigates the relationship between weight sparsity and the robustness of the models against adversarial attacks. Other provable defenses utilize k-nearest neighbor (KNN) \cite{wang2018analyzing, papernot2018deep}, and Bayesian deep neural network (BNN) \cite{liu2018adv}.

Ensemble methods \cite{tramer2018ensemble, strauss2017ensemble, chen2020adversarial} played a big influence in this work. However, we must emphasize that this is not an ensemble method. It is true that multiple submodels are involved in the creation of this defense. Nevertheless, only one submodel will answer a given classification query and the final output will not be based on the consensus of the submodels. The purpose of not following the traditional ensemble approach is to present to the attacker a different target model at each interaction limited only by the number of diverse submodels in the population.
Though the environment is white-box, the attacker if forced to keep guessing which model will be facing on each attack.

The methodology introduced in this work builds on the work of Zhou, Kantarcioglu, and Xi \cite{zhou2018breaking} where we adapt the element of randomness as a defense mechanism against adversarial attacks. The work of \cite{wang2020advms} follows the same approach but with adversarially trained models. The work of \cite{wang2019protecting} utilizes several random switching blocks to "prevent adversaries from exploiting fixed model structures."

\section{Preliminaries}
The experiments in this work are based on two types of deep learning network populations.
A population consists of $n$ submodels (networks) but operates as a single model.
The first type consists of a population of LeCun's et. al. \cite{lecun1998gradient} LeNet5 submodels trained on the MNIST dataset. 
Each of these submodels consists of three convolutional layers followed by a fully connected layer and achieves an accuracy of 99.2\% on benign data. 
LeNet5 is small enough which would help us to easily examine and investigate the weights produced and explain the rationale behind the results achieved. 
The second type consists of a population of ResNet18 submodels \cite{he2016deep} trained on the CIFAR-10 dataset, where they achieve an accuracy of 95.4\% on benign data. 
The models are tested against multiple attacks.
The focus of this work is on the FGSM, PGD, AutoAttack \cite{croce2020reliable}, MI-FGSM \cite{dong2018boosting}, and the Expectation Over Transformation (EOT) attack \cite{athalye2018synthesizing}.
FGSM and PGD are known stable attacks in the literature.
AutoAttack is a strong attack that we use to test this methodology to its limits.
MI-FGSM attack enhances the transferability of the crafted attacks which is suitable to counter our methodology since the defending submodel is different every time an attack is conducted.
The EOT attack takes the gradient generated, samples it for $n$ times and considers the average of all the samples to build the attack on.
This attack is interesting to us since it is created for the purpose of countering defenses that are dependent on the element of randomness.
The adversarial attacks Python library used in this work is the Torchattack library \cite{kim2020torchattacks}. The source code of this library can be accessed on GitHub \footnote{\url{https://github.com/Harry24k/adversarial-attacks-pytorch}} 

\begin{figure}[t]
  \centering
  \includegraphics[width=0.45\linewidth]{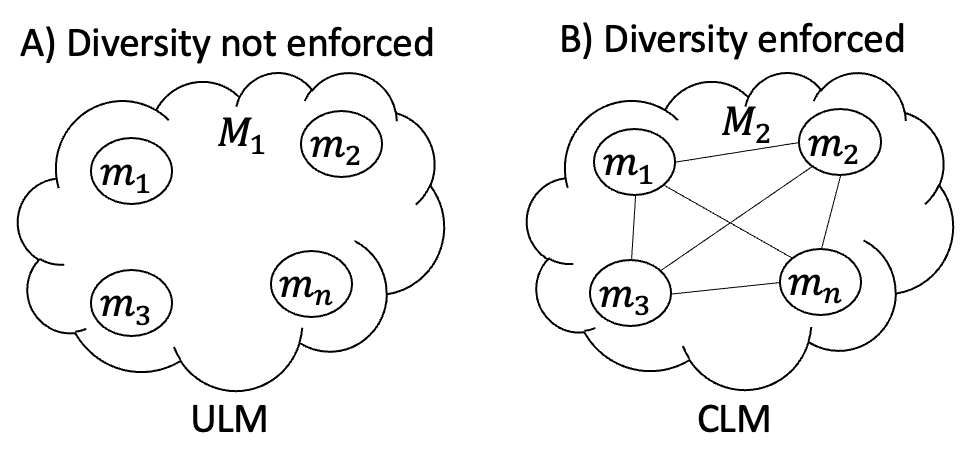} 
  \caption{A) Model $M_1$ consists of a population of $n$ submodels trained to accomplish task at hand with no diversity enforcement mechanism (ULM). B) Model $M_2$ consists of a population of $n$ submodels trained with a diversity enforcement mechanism: counter linking weights (CLM).} 
  \label{fig:ulm_clm} 
\end{figure}

The experiments compare the performance of these population-based models operating under two different training regimes. We name them Unlinked Model (ULM) and CLM as shown in Fig.\ref{fig:ulm_clm}.
ULM consists of a population of independently trained submodels without enforcing any diversity or conducting any other form of interference during their training.
This methodology is based on previous work \cite{zhou2018breaking, wang2020advms} to break the transferability of the adversarial attack.
CLM consist of a population of submodels trained using a mechanism to enforce submodel diversity.

There are two main hyper-parameters for the CLM training, $\alpha$ and $\delta$.  
For the LeNet5 submodels, $\alpha$ and $\delta$ are both set to $0.1$.  
While in the ResNet18 submodels, $\alpha$ is set to $0.5$ and $\delta$ is set to $0.1$. 
When training LeNet5 submodels with adversarial samples, $\alpha$ is set to $0.01$ and $\delta$ to $5e-3$, and when training ResNet18 submodels with adversarial samples, $\alpha$ is set to $0.1$ and $\delta$ to $0.05$.
These hyper-parameter values provided the best results which are shown in section(\ref{sec:exp_disc}).

Our code for this work can be accessed using the following link\footnote{\url{https://drive.google.com/drive/folders/1Dkupi4bObIKofjKZOwOG0owsBFwfwo_5?usp=sharing}}

\section{Methodology} 
\label{sec:method}

In this work, we introduce a model that consists of diverse submodels where a submodel is selected at random for every classification query.  
The attacker has no practical way of knowing which submodel will do the classification next, same as the defender. However, the attacker has full knowledge of the submodels' parameters.  
The attacker has a non trivial task of figuring out which submodel to interact with next.
This creates a model that is robust against adversarial attacks.
The reason is that deep learning models are stochastic algorithms where the random weight initialization allows the submodel to learn from a different starting point in the search space. 
Therefore, the gradients produced by these submodels become more orthogonally similar which would cause the robustness in the system.

\begin{figure}[t]
  \centering
  \includegraphics[width=0.5\linewidth]{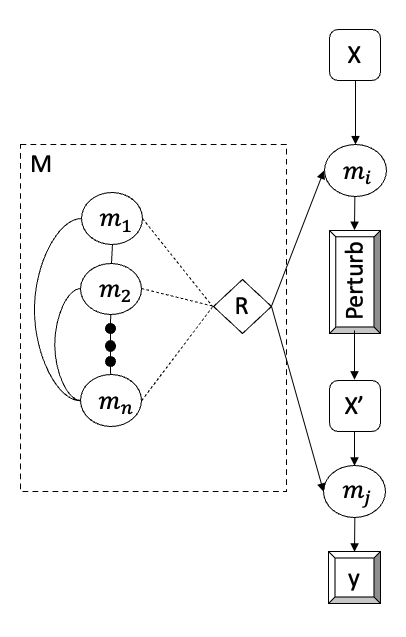} 
  \caption{The attack flow: on model $M$, which consists of $n$ submodels, a benign sample $x$ is used to craft an attack with one of $M$'s submodels $m_i$ selected by the $Random$ function. The adversarial sample $x'$ is sent to $M$ again for classification. $M$ dispatches a different submodel $m_j$, also selected by the function $Random$ with a high probability that $m_j \neq m_i$.} 
  \label{fig:attack_flow} 
\end{figure}

Consider a model comprising of a population of $n$ submodels $\{m_1, m_2, m_3, ..., m_n\} \in M$ so that every time there is an image classification query, a submodel selected at random $m_i \in M$ is dispatched for the job.
The function $Random$ facilitate the selection of $m_i$ over a uniform distribution.
Fig.\ref{fig:attack_flow} shows the process of calculating the gradient of the image to perturb it using submodel $m_i$, selected at random, then classifying it with another submodel $m_j$ where there is a high probability that $m_i \neq m_j$.

An attack crafted using submodel $m_i$ would have to be executed on the same submodel ($m_i$) to guarantee a high probability of success.
However, since the submodel selection for image classification is a random process, the attack might be executed on a different submodel $m_j$, decreasing the probability of the attack's success.
This creates a non-trivial task for the attacker to distinguish which submodel being interacted with at each step.

\begin{figure}[h]
  \centering
    \includegraphics[width=0.24\linewidth]{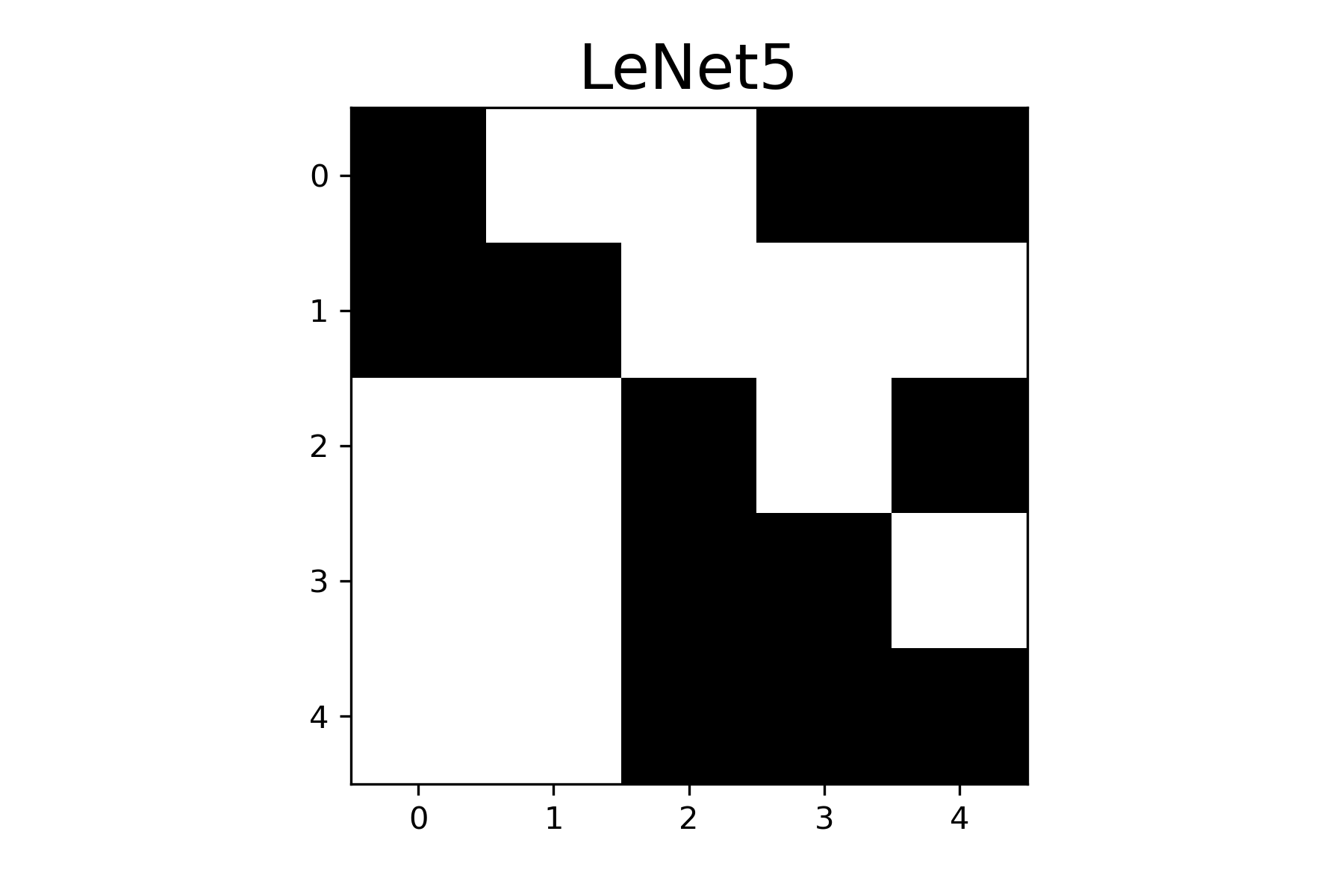}
    \includegraphics[width=0.24\linewidth]{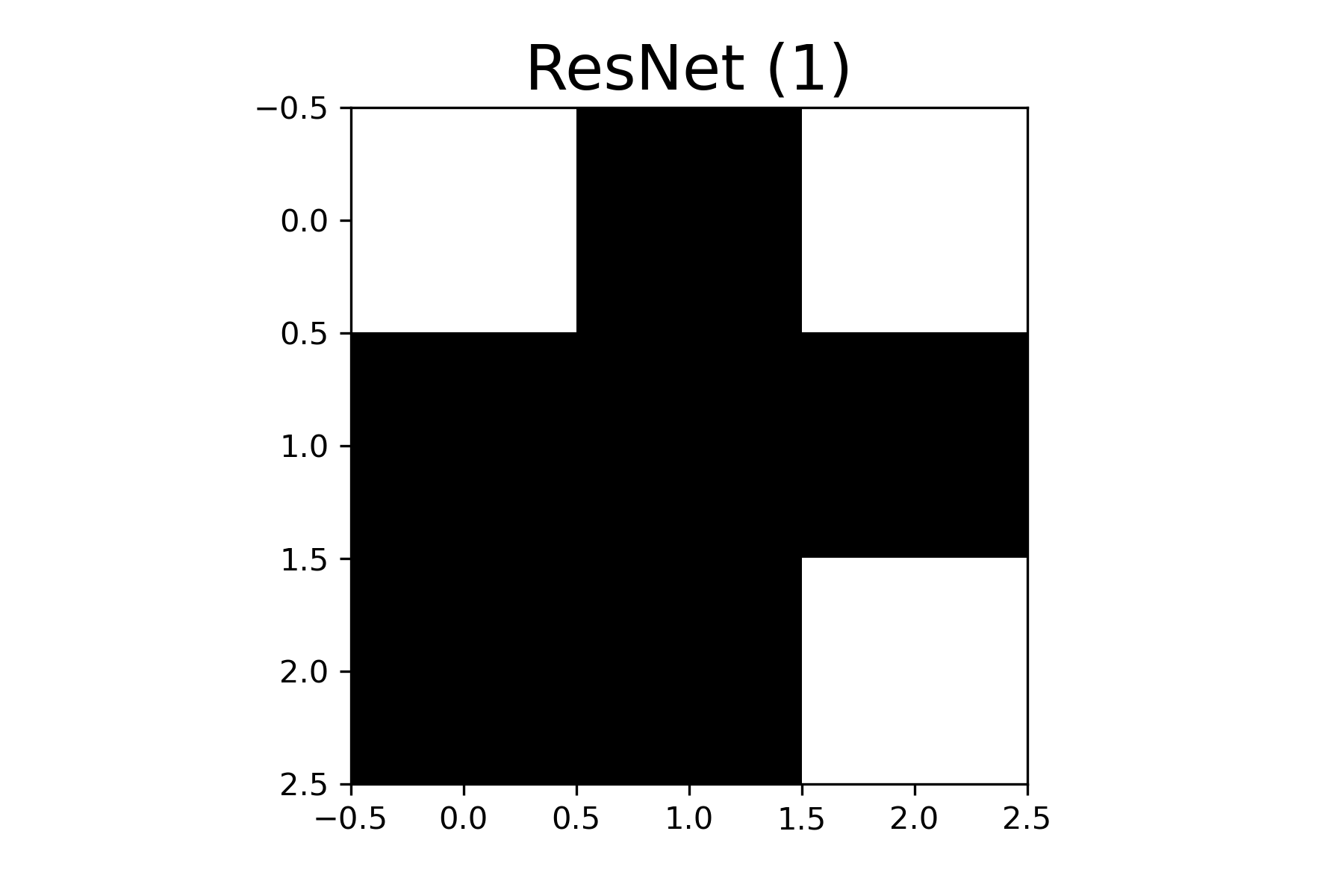}
    \includegraphics[width=0.24\linewidth]{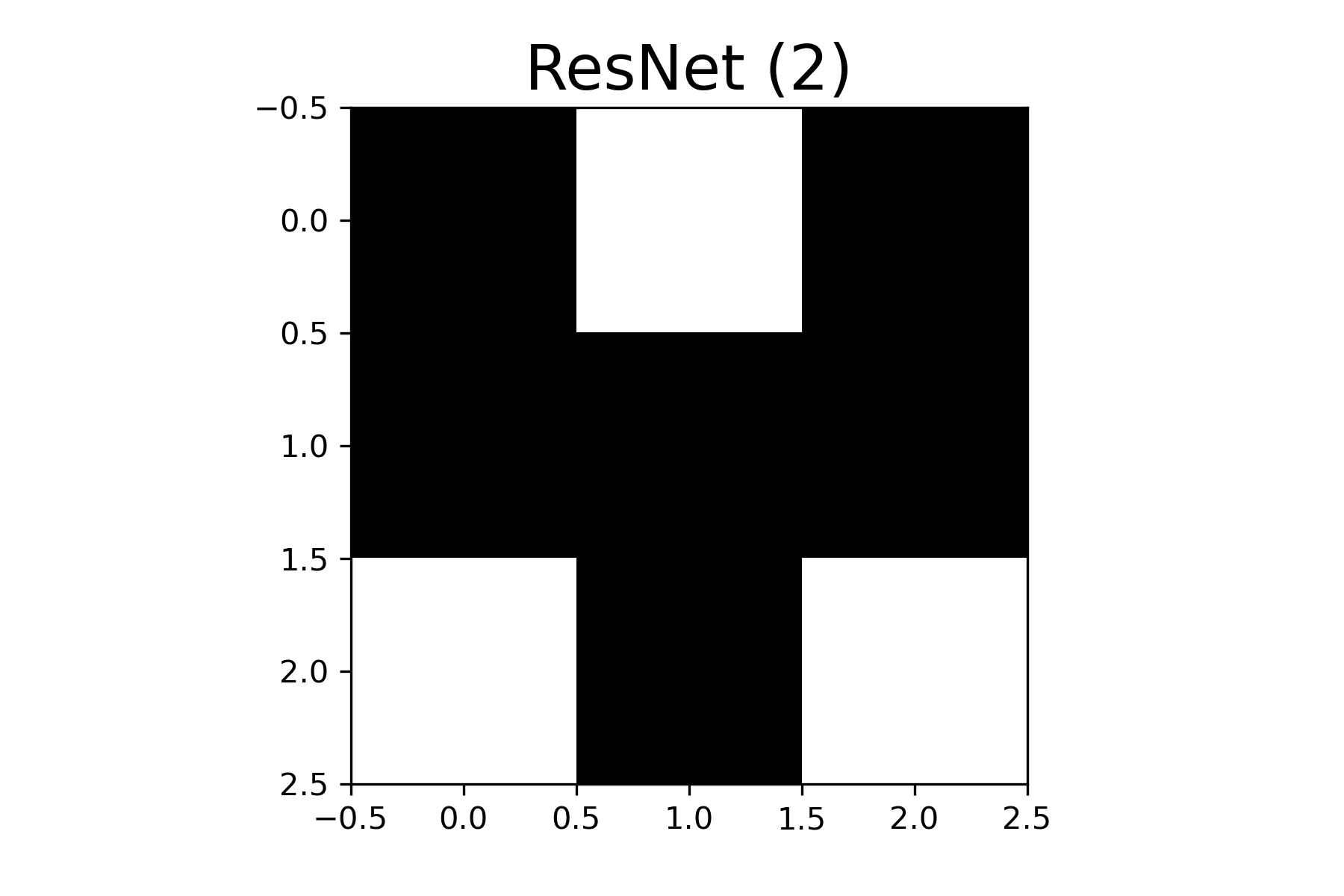}
    \includegraphics[width=0.24\linewidth]{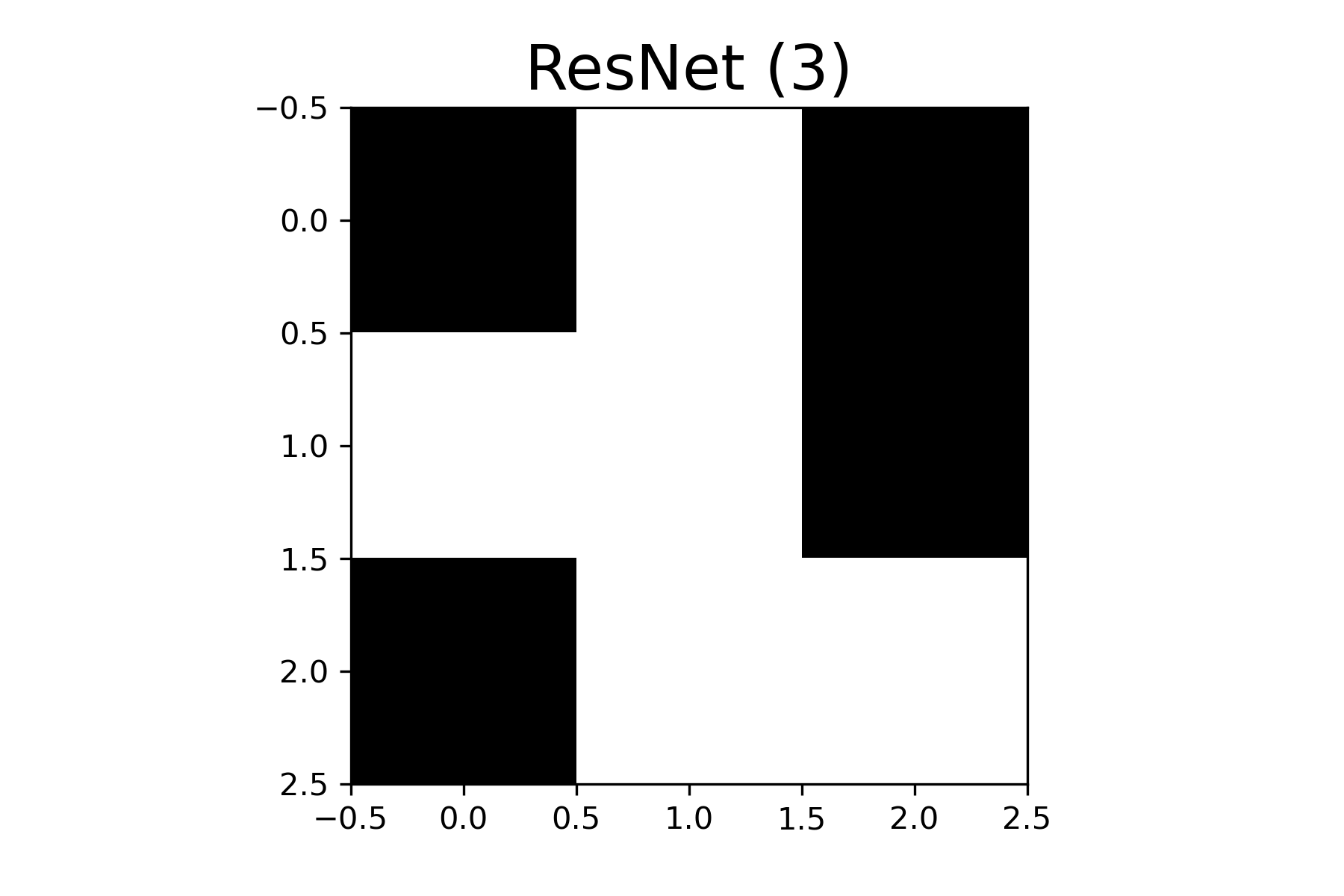}
    
  \caption{The Masks: the LeNet5 subfigure shows a 5 by 5 mask. ResNet(1-3) subfigures shows a 3 by 3 masks.}
  \label{fig:masks}
\end{figure}

All the submodels share the same architecture which allows us to monitor them and force their weights to be different through the process of training or retraining them with counter-linking.
$\{w,w_c\} \in W^l_i$ are two types of weights in the weight matrix of layer $l$ where $W_i = \{W^1_i, W^2_i, ..., W^L_i\}$ is a set of all weights in a submodel $W_i$. $w$ is a normal weight while $w_c$ is a counter linked weight (CLW).
All $w_c$ weights are determined by a uniform distribution over $W^l_i$ meaning that around half of the weights in layer $l$ will be $w_c$. In this work, $l$ is always set to 1. In other words, we always counter link the first weight layer of the submodels.
If we have a $w_c$ in one of the submodel's $W^l$, then all submodels would have a coinciding weight that is also $w_c$ in the same weight matrix of layer $l$.
Fig.\ref{fig:masks} shows the randomly generated link tensors $\ell$ that defines the indices of the weights that will be labeled as $w_c$.
These tensors will be referred to as masks from now on. 
Notice that in the LeNet5 submodels, the filter size is 5x5 kernel with 1-channel input while in the ResNet18 submodels the filter size is 3x3 kernel with 3-channel input.
There are 6 filters in the first layer of convolution of LeNet5 and so the mask is applied to the 6 feature maps when compared with another submodel during the retraining of the submodels.
Same for the ResNet submodels, however there are 64 filters in there.

The training of the CLM is simultaneous where every submodel is trained for one epoch until all the submodels in the population finish that epoch, then the training is moved to the next epoch.  
During the training, a periodic similarity examination (PRSE) step is conducted to measure the weight similarity between the model currently being trained $m_i$ and another model $m_j$ selected at random to assure diversity in their weights such that $m_i \neq m_j$.
If the difference between $w_c$ from $m_i$ and $m_j$ is less than the threshold $\delta$, then a small amount of noise $\alpha$ is added to $\forall w_c$ in $m_i$.  
 
Alg.\ref{alg:model_training} explains how the submodels are trained with this methodology.
If the submodel being trained in the current epoch has been trained for $n=10$ steps, then the function $Random$ would randomly selected a submodel from $\{m_1, m_2, ..., m_n\}$ such that the submodel selected at random is not the same one being trained, and the function PRSE is executed with parameters $m_i$ (The submodel currently under training) and $m_j$ (The submodel selected by the function $Random$). 

\begin{algorithm}
\caption{Model Training Loop}
\label{alg:model_training}
\begin{algorithmic}
    \State \texttt{$n = 10$}
    \For{\texttt{epoch in epochs}}
        \State \texttt{$c = 0$}
        \For {\texttt{$m_i$ in $\{m_1, m_2, ..., m_n\}$}}
            \If {$c > n - 1$}
                \State \texttt{$m_j = Random(\{m_1,m_2,...,m_n\} - \{m_i\})$}
                \State \texttt{$PRSE(m_i, m_j)$}
                \State \texttt{$c = 0$}
            \EndIf
            \State \texttt{$Train(m_i)$}
            \State \texttt{$c++$}
        \EndFor

    \EndFor
\end{algorithmic}
\end{algorithm}

In PRSE, $\alpha$ is added to $w_c$ in $m_i$.
Nevertheless, we need to define which $w_c$ absolute difference between submodels $m_i$ and $m_j$ is below the threshold $\delta$.
Equation(\ref{eq:beta}) labels the indices $i$ of $W_i^l$ that satisfies the threshold $\delta$.

\begin{equation} 
    \beta =  
    \begin{cases} 
      1 & \text{if $|w_{i_i}^l - w_{i_j}^l| < \delta$ \;\;\;\;\;\; $w_{i_i}^l \in W_i^l, \;\; w_{i_j}^l \in W_j^l$ }\\ 
      0 & \text{otherwise} 
    \end{cases}  
    \label{eq:beta} 
\end{equation} 

After marking the indices of those $w$ which satisfied the threshold $\delta$, only those among them who are $w_c$ are the ones who $\alpha$ will be added to.
This can be achieved using the masks discussed above in Fig.\ref{fig:masks}.
To add $\alpha$ only to $w_c$ in $W^l_i$, we start with a zero tensor $O$ that has the same dimensionality as $W^l_i$. 
A negative or positive $\alpha$ is added to it, depending on the index of the submodel $n$ where the $w_c$ of a submodel with an odd index will always be diverged in a negative direction and positively otherwise.
An element-wise product between the positive/negative values of the $O$ tensor with $\beta$ would keep the values of the $\alpha$s that need to be added to $w$ that satisfies the threshold $\delta$ while zeroing the rest of the tensor values.
Then to be followed by another element-wise product with the tensor $\ell$ (the mask) which would keep the values of $\alpha$s that need to be added to $w_c$ and zeroing the rest.
Finally, The values of the current $W^l_i$ will be added so that $\alpha$ will only be accumulated to $w_c$ that satisfies the threshold $\delta$ and keep the same values for the rest. 
Equation(\ref{eq:weight}) shows how a submodel's weights are updated to maintain diversity from the other submodels. The new weight matrix of layer $l$ at step $t+1$ is the result of adding the $\alpha$ noise to $\forall w_c \in W^l_i$ at step $t$ that satisfies the threshold $\delta$.

\begin{equation} 
    W^{'l}_{i_{t+1}} = (O + (-1)^n * \alpha) \odot \beta \odot \ell + W_{i_{t}}^l 
    \label{eq:weight} 
\end{equation}

\begin{figure}[t]
  \centering
    \includegraphics[width=0.45\linewidth]{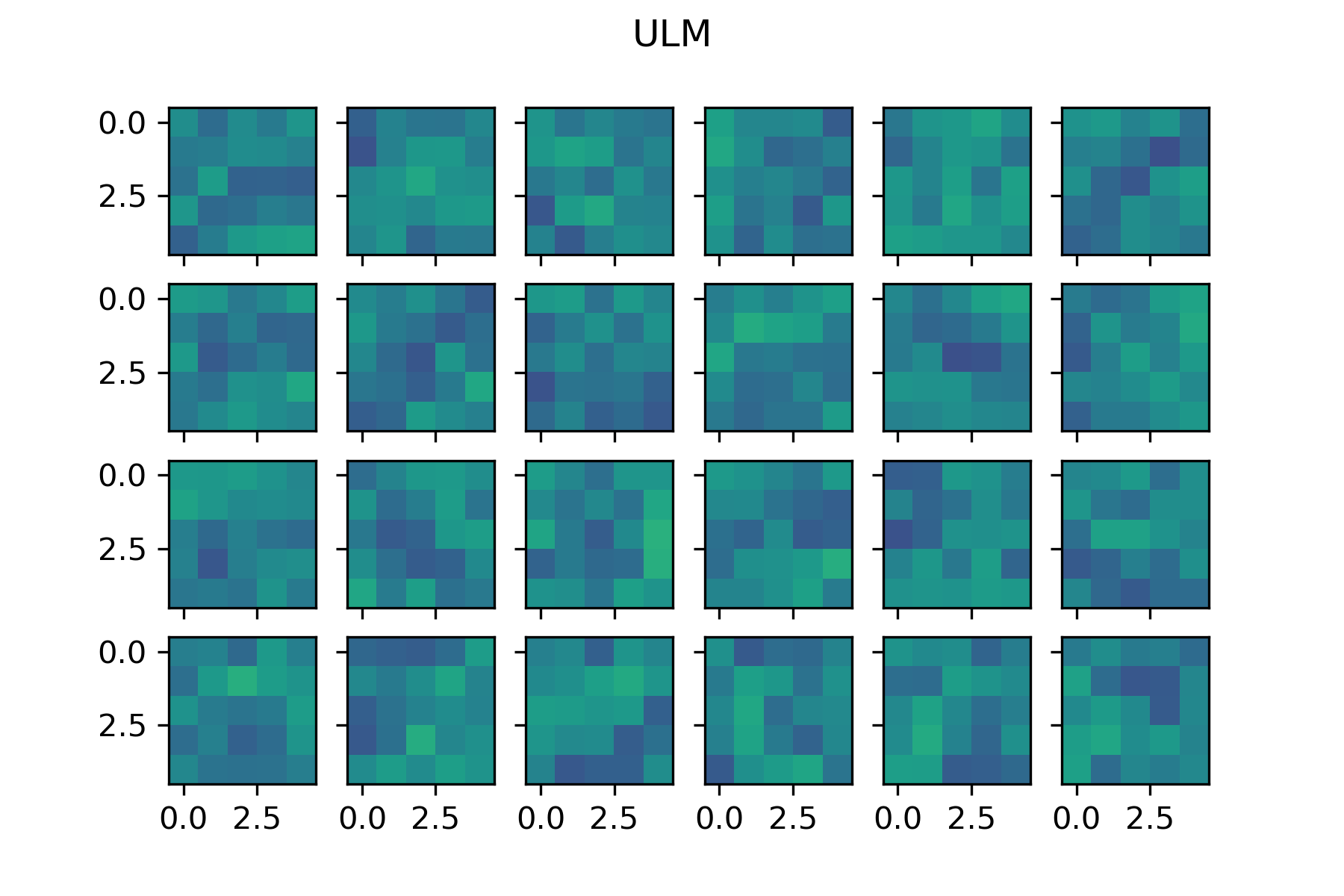} 
    \includegraphics[width=0.45\linewidth]{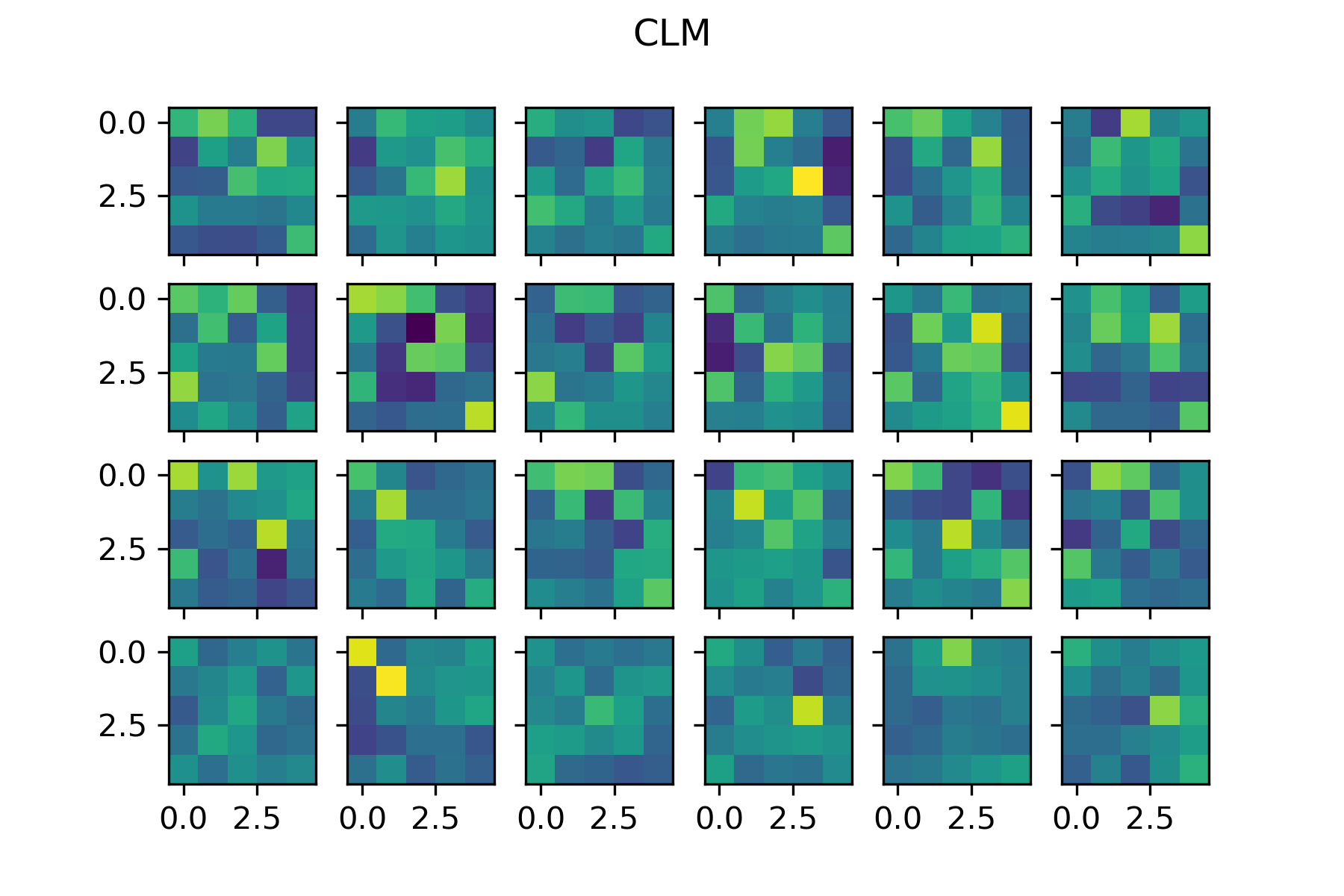}
  \caption{The Counter-Linked Model (CLM) weights have a higher standard deviation in their distribution than the smoother Unlinked Model (ULM) weights. This shows that the counter-linking forced $w_c$ in CLM to become different and created diversity in the submodels weights.}
  \label{fig:conv1}
\end{figure}

Fig.{\ref{fig:conv1}} shows the variety in weights between 4 submodels in a CLM and corresponding 4 in the ULM.
Each row represents the 6 filters of either a CLM submodel or an ULM submodel.
Notice how the weights within the CLM are more diverse in values than the ULM.
They have higher standard deviation in their distribution than the smoother corresponding ones in ULMs.

\section{The defense environment and measuring its robustness} 

In this work, the environment is a white-box setting.
The attacker has full knowledge of the model, its submodels' architecture, weights, and the uniform distribution of $Random$ but lacks the knowledge of test-time randomness of when a submodel is selected for the classification query.
Same as the defender. 
The flow of the attack scenario (Fig.\ref{fig:attack_flow}) is as follows:  
\begin{enumerate}
    \item the attacker would use a submodel at random to craft an attack using an image.  
    \item The attacker queries the perturbed image for classification.
    \item The system dispatches a submodel randomly selected for the job. 
\end{enumerate}
  
Obviously, the more submodels in the population, the lower the probability that the same submodel used for crafting the attack will be used for classification.  
If only two submodels are in the population, then the attacker has a $50\%$ chance of targeting the same model used for crafting the attack.  
The probability of targeting the right submodel is represented as the linear function $\frac{1}{n}$ where it converges to 0 rapidly.  
If the population holds 10 submodels, then the probability for targeting the right model by the attacker is $0.1$ every time the attacker plans an attack.  
This increases the probability of a likely successful defense whenever an attack is conducted. 
We noticed that adding more submodels would not affect the robustness of the model coinciding with the conclusion from the work of \cite{zhou2018breaking}. Therefore we set our population to 10 submodels.

For a reliable robustness measure, the model robustness average (MRA) is considered.
MRA is the robustness average of all submodels in the population such that every submodel's robustness is measured on the entire perturbed dataset and then the average of all these individual robustness is calculated.
 Equation.\ref{eq:mra} explains how the MRA is calculated.
 $x' \in X'$ which is set of perturbed images where each sample is perturbed by a random submodel $m_i \in M$. The function $\rho$ predicts every sample in set $X'$ and return the ratio of correct prediction over the total number of samples $N$ (The robustness accuracy of model $m_i$). The process is then repeated for all submodels in model $M$ and the average of these accuracies is then taken. 

\begin{equation}
    MRA = \frac{1}{n} \sum_{i=1}^{n}\left(\frac{1}{N} \sum_{j=1}^{N} \rho(m_i, x'_j)\right)
    \label{eq:mra}
\end{equation}

\section{Experiment} 
\label{sec:exp_disc}
  
The first step of action is to check whether an accuracy drop occurred after retraining the ULM with counter-linking. 
No drop of accuracy or at least a minor drop is expected if occurred.
However, since adversarially training a model would cause an accuracy drop, then we expect that the CLM-Adv to also be affected by it.
Table(\ref{tab:benign_acc}) shows the MRAs on both MNIST and CIFAR-10 datasets benign samples.

\begin{table}
\caption{MRAs on benign samples of both datasets.}
  \centering
  \begin{tabular}{|c|c|c|}
    \hline
     \textbf{Model\textbackslash Attack}& \textbf{MNIST} & \textbf{CIFAR-10} \\
    \hline
    \textbf{ULM} &  99.04\% & 95.34\% \\
    \textbf{CLM} &  98.54\% & 95.24\% \\
    \textbf{ULM-Adv} & 98.14\% & 84.37\% \\
    \textbf{CLM-Adv} & 98.18\% & 83.57\% \\
    \hline
  \end{tabular}
  \label{tab:benign_acc}
\end{table}

Indeed, the noise $\alpha$ added to the submodels' weights in CLM during their training does not cause a noticeable accuracy drop. 
The submodels in CLM maintained their accuracy of $98.5\% - 99.04\%$ for LeNet5 and $95.2\%$ for ResNet18. 
We also did a conformity test where we trained 5 CLMs, and the accuracy range is also maintained in these 5 models for both experiments. 

\begin{figure}[t]
  \centering
    \includegraphics[width=0.45\linewidth]{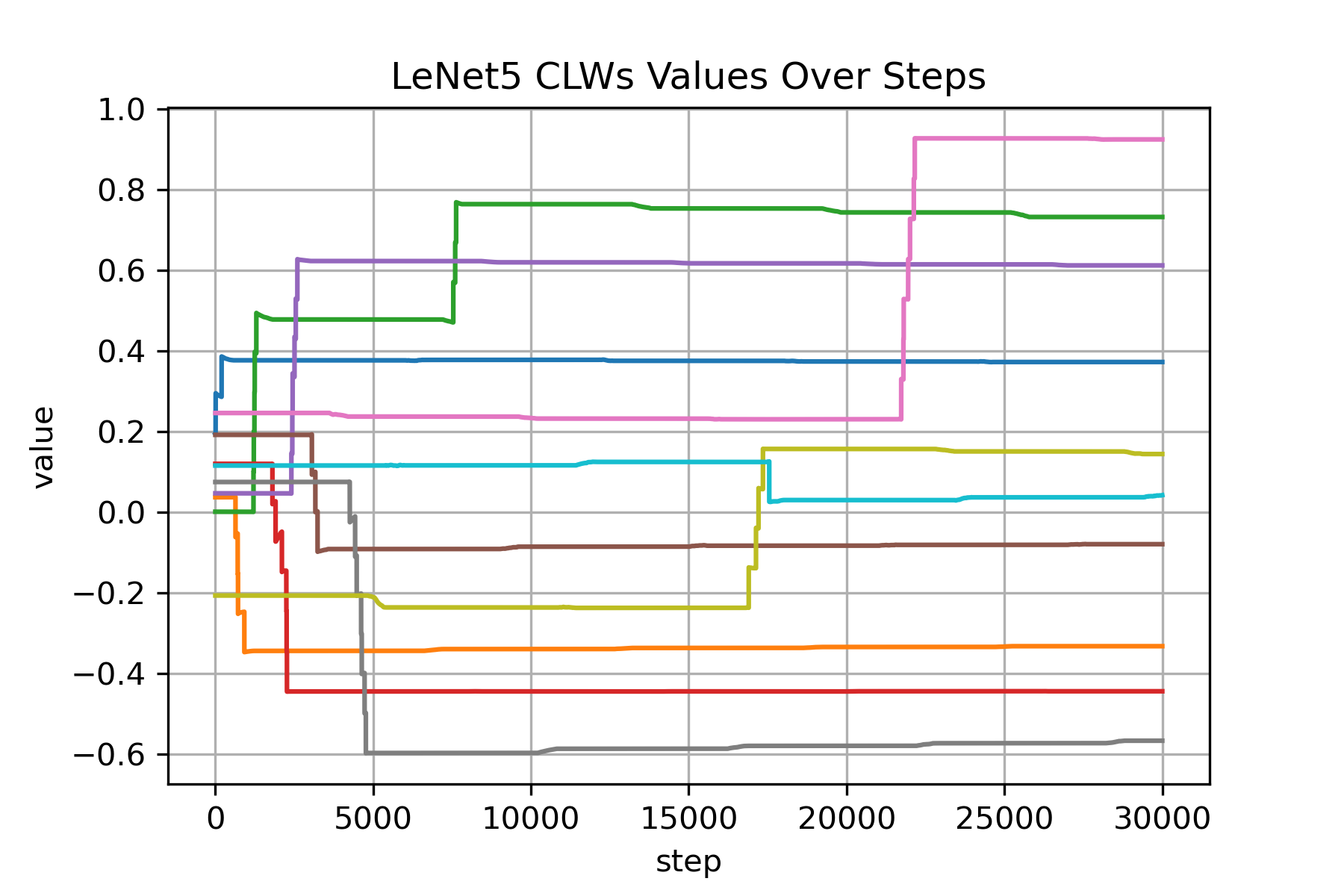} 
    \includegraphics[width=0.45\linewidth]{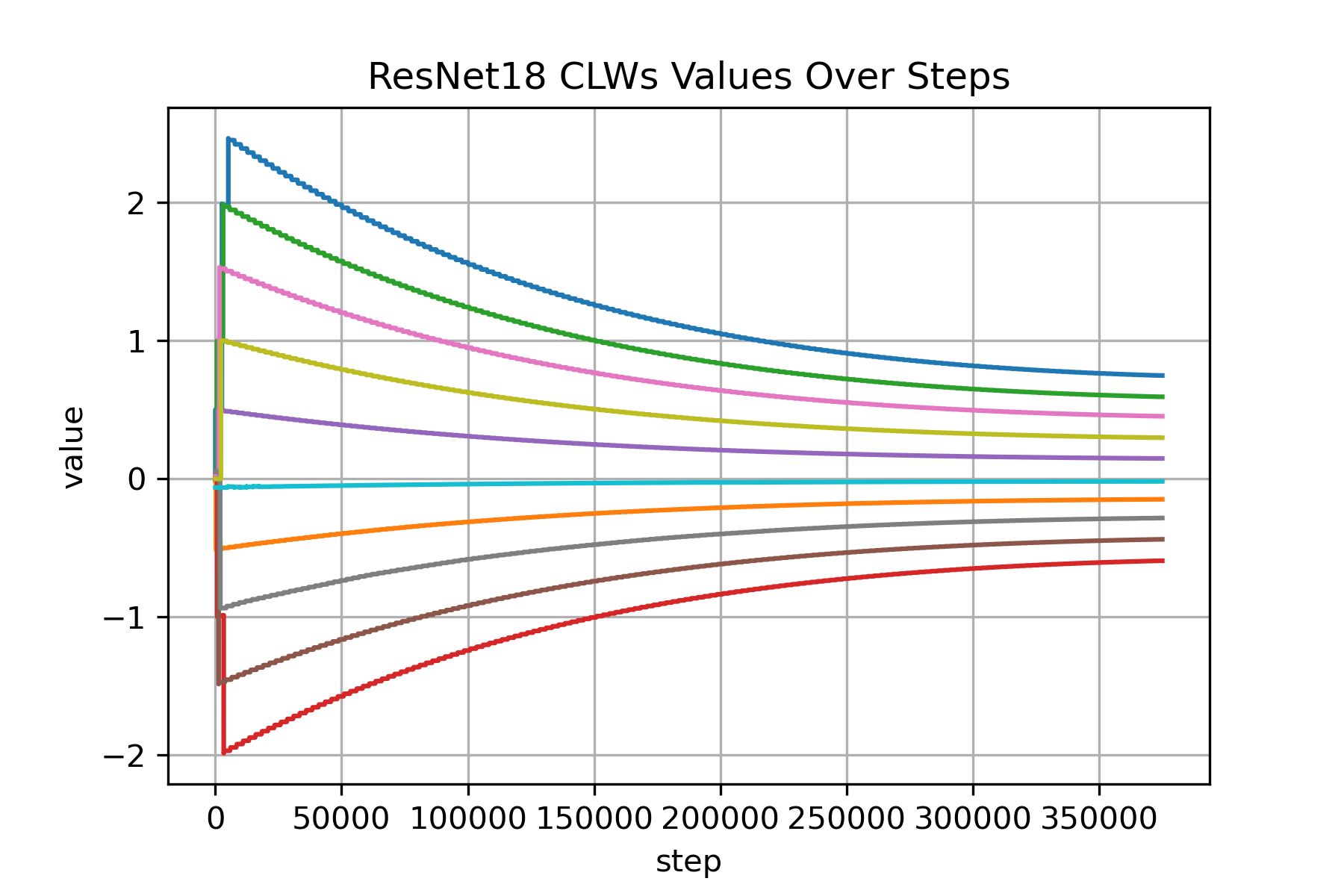}
  \caption{The monitored CLWs that coincide in all submodels in both type of networks, LeNet5 and ResNet18. In LeNet5 submodels, the CLWs kept diverging until all converged on their local minimum while the CLWs in the ResNet18 submodels only diverged at the very beginning and slowly converged to different local minimums.}
  \label{fig:cln_values}
\end{figure}

The purpose of retraining an ULM with counter-linking, thus becoming a CLM, is to conduct a sanity check on this methodology.
To show that the MRA would increase when ULM is retrained into a CLM.
CLM can be trained from scratch, without retraining it from an ULM.
However, the hyper-parameters $\alpha$ and $\delta$ would have to be adjusted depending on the model, the learning rate won't be decreased as we explain below, and more epochs are needed to fully train the submodels.
In the retraining process, the learning rate is decreased so that the retrained submodels won't rapidly converge to the same local minimum they converged to when trained as an ULM.
For the LeNet5 submodels, the learning rate was set to $1e-4$, and for the ResNet18 submodels, it was set to $1e-2$.
LeNet5 submodels are retrained for 5 epochs and the ResNet18 submodels are retrained for 150 epochs.
To show how the coinciding CLWs in all submodels diverge during the retraining, a single weight and all its coinciding CLWs in all submodels is monitored to verify the divergence.
Fig.\ref{fig:cln_values} shows the monitored coinciding CLWs over the 10 submodels in both networks types (LeNet5, ResNet18) retraining.
Notice that some CLWs' weight are diverged in a positive directions while others in a negative directions as explained in section(\ref{sec:method}), equation(\ref{eq:weight}).
The CLWs of the LeNet5 submodels continued diverging throughout the retraining period while the CLWs of the ResNet18 submodels diverged only at the early steps of the retraining process and converged slowly to different minimums.

\subsection{\textbf{Results}}

ULM and CLM were also trained with adversarial samples (ULM-Adv, CLM-Adv) which achieved a state-of-the-art robustness. 
The adversarial samples were generated using the PGD attack.
On the MNIST dataset, the step size is $0.01$ over $100$ steps with perturbation magnitude $\epsilon=0.3$.
While on the CIFAR-10 dataset, the step size is $2/255$ over $20$ steps with $\epsilon=8/255$.
  
Considering $\epsilon=0.3$ when attacking the models with the FGSM attack, the ULM achieved an MRA of $27.68\%$ and after retraining the model with counter-linking, the MRA increased to $51.36\%$. 
Under the PGD attack, CLM increased the MRA by around $20$ points.
CLM even performed better under the AutoAttack, and MI-FGSM attacks.
It performed best under the EOT attack.
ULM-Adv and CLM-Adv performed similarly under all attacks variations.
Table(\ref{tab:acc_mnist}) shows the MRA values for all models.

\begin{table}[htbp]
  \caption{MRA of the models against different adversarial attacks with $\epsilon=0.3$ on the MNIST dataset}
  \label{tab:acc_mnist}
\begin{center}
\begin{tabular}{|c|c|c||c|c|}
\hline
     \textbf{Attack\textbackslash Model}& \textbf{ULM} & \textbf{CLM} & \textbf{ULM-Adv} & \textbf{CLM-Adv} \\
\hline
    \textbf{FGSM} &  27.68\% & 51.36\% & 94.35\% & 94.34\%\\

    \textbf{PGD} &  35.94\% & 54.29\% & 91.30\% & 91.00\% \\

    \textbf{AutoAttack} & 22.80\% &  45.43\% & 84.32\% & 84.43\% \\

    \textbf{MI-FGSM} & 29.00\% & 50.04\% & 91.19\% & 90.51\% \\
    
    \textbf{PGD-EOT} & 40.46\% & 58.20\% & 92.83\% & 92.44\% \\
\hline
  \end{tabular}
\end{center}
\end{table}

ULM performed well against the FGSM attack on the CIFAR-10 dataset. 
However, their robustness faltered when attacked with the PGD, AutoAttack, MI-FGSM, and EOT attacks.
The perturbation magnitude $\epsilon$ used is $8/255$.
CLM performed better than ULM against all attacks.
Achieving an MRA of $54.04\%$ against the FGSM attack which is slightly better than ULM while doubling the value against the PGD AutoAttack, and EOT attacks.
MI-FGSM affected CLM on this dataset which managed an MRA of $22.13\%$, unlike the MRA achieved on the MNIST dataset.
Nevertheless, still better than the MRA achieved by the ULM.
Similar to the results above, ULM-Adv and CLM-Adv performed similarly against these attacks on this dataset.
Table(\ref{tab:acc_cifar}) shows the MRA values for all models.

\begin{table}
  \caption{MRA of the models against different adversarial attacks on the CIFAR-10 dataset}
  \label{tab:acc_cifar}
  \centering
  \begin{tabular}{|c|c|c||c|c|}
    \hline
     \textbf{Attack\textbackslash Model} & \textbf{ULM} & \textbf{CLM} & \textbf{ULM-Adv} & \textbf{CLM-Adv} \\
    \hline
    \textbf{FGSM} &  50.79\% & 54.04\% & 62.91\% & 62.97\%\\
    \textbf{PGD} &  12.36\% & 35.84\% & 58.85\% & 59.16\% \\
    \textbf{AutoAttack} & 12.23\% &  39.76\% & 61.70\% & 61.61\% \\
    \textbf{MI-FGSM} & 15.21\% & 22.13\% & 61.50\% & 61.58\% \\
    \textbf{PGD-EOT} & 12.50\% & 35.16\% & 59.30\% & 59.45\% \\
    \hline
  \end{tabular}
\end{table}

To verify that ULM-Adv and CLM-Adv doesn't perform the same against all the attacks considered in this work, the robustness of these two models are tested with a higher perturbation magnitude.
An $\epsilon$ of 0.4 considered for the MNIST dataset, and $16/255$ for the CIFAR-10 dataset.
On MNIST, CLM-Adv performed better against FGSM based attacks (FGSM and MI-FGSM) while ULM-Adv performed better against PGD based (PGD, AutoAttack, and PGD-EOT).
CLM-Adv performed slightly better against all attacks on CIFAR-10.
Table(\ref{tab:adv-mnist}) and Table(\ref{tab:adv-cifar}) shows the results of the ULM-Adv and CLM-Adv against adversarial attacks with higher $\epsilon$ on both datasets.

\begin{table}
  \caption{MRA of adversarially trained models against adversarial attack ($\epsilon=0.4$) on MNIST}
  \label{tab:adv-mnist}
  \centering
  \begin{tabular}{|c|c|c|}
    \hline
     \textbf{Attack\textbackslash Model} & \textbf{ULM-Adv} & \textbf{CLM-Adv} \\
    \hline
    \textbf{FGSM} &  72.24\% & 75.31\%\\
    \textbf{PGD} &  39.54\% & 32.74\%\\
    \textbf{AutoAttack} & 13.92\% &  11.75\%\\
    \textbf{MI-FGSM} & 67.49\% & 70.01\%\\
    \textbf{PGD-EOT} & 54.12\% & 47.48\% \\
    \hline
  \end{tabular}
\end{table}

\begin{table}
  \caption{MRA of adversarially trained models against adversarial attack ($\epsilon=16/255$) on CIFAR-10}
  \label{tab:adv-cifar}
  \begin{center}
  \begin{tabular}{|c|c|c|}
    \hline
     \textbf{Attack\textbackslash Model} & \textbf{ULM-Adv} & \textbf{CLM-Adv} \\
    \hline
    \textbf{FGSM} &  45.41\% & 46.32\%\\
    \textbf{PGD} &  30.97\% & 32.58\%\\
    \textbf{AutoAttack} & 31.20\% &  32.84\%\\
    \textbf{MI-FGSM} & 39.83\% & 41.20\%\\
    \textbf{PGD-EOT} & 31.46\% & 33.15\% \\
    \hline
  \end{tabular}
  \end{center}
\end{table}
  
\subsection{\textbf{Discussion}}
The question that arises after this experiment is why the MRA of the CLM is higher than the ULM?
Previous work \cite{pang2018towards, pang2019improving, zhou2021exploring} explained that promoting diversity in an ensemble to make it difficult for adversarial attacks to transfer among the individual networks. The work of \cite{pang2019improving} promotes diversity among the networks by introducing the Adaptive Diversity Promoting (ADP) regularizer which "encourages the non-maximal predictions of each member in the ensemble to be mutually orthogonal, while keeping the maximal prediction be consistent with the true label."
The work of \cite{bogun2021saliency} diversifies the ensemble by promoting saliency map diversity. In our work, the counter linking mechanism promotes saliency map diversity. 
Therefore, attempting to understanding this, the saliency maps produced by both model types are investigated.
The cosine similarity (equation \ref{eq:cos}) is measured between every input (from the test dataset) for every two submodels from the same population, e.g., submodels 1 and 2 from the CLM population.
For $n$ number of submodels, we have $(n^2 - n) / 2 $ number of comparisons.
In this work's case, 10 submodels yields 45 comparisons.
$s_{m_i}$ is the saliency map generated by model $i$ and $s_{m_j}$ is the saliency map generated by model $j$. In equation \ref{eq:sm}, the saliency map is calculated by the partial derivative of the submodel output $y$ with respect to the input $x$.
  
\begin{equation} 
    cos\_sim(s_{m_i},s_{m_j}) = \frac{s_{m_i} \cdot s_{m_j}}{\lVert s_{m_i} \rVert \lVert s_{m_j} \rVert} 
    \label{eq:cos} 
\end{equation} 

\begin{equation}
    s_m = \frac{\partial f_m(x)[y]}{\partial x}
    \label{eq:sm}
\end{equation}

\begin{figure}[t] 
  \centering 
  \includegraphics[width=0.4\linewidth]{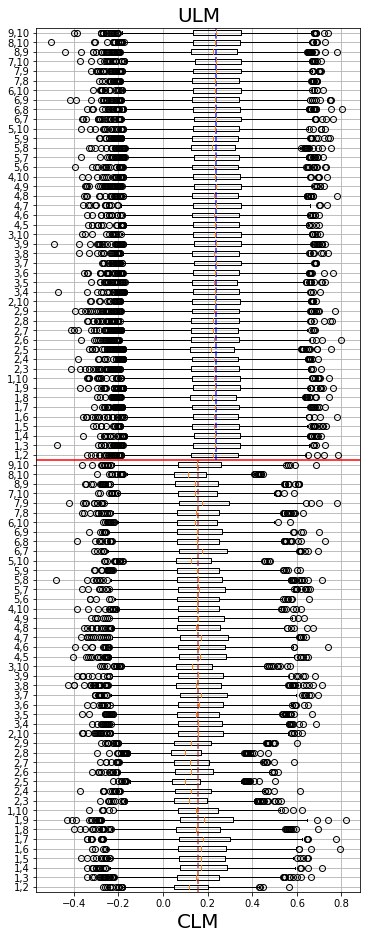} 
  \caption{Gradient similarity between every pair of submodels in both CLM and ULM. Below of the red line are the gradient similarities for the CLM. Above are the similarities for the ULM. Notice that the medians of the gradients similarities between the submodels of the CLM are closer to 0 compared to the medians from the ULM.} 
  \label{fig:box} 
\end{figure}

In Fig.\ref{fig:box}, the saliency map similarity between every two submodels of the same population is shown for both CLM and ULM.
This was conducted with the ResNet18 models on the CIFAR-10 dataset.
Every similarity comparison is shown as a box plot that provides the range of saliency maps similarities between every two submodels. 
There are 90 similarity comparisons in total.
The red horizontal line is a separator between the results.
Above it are the ULM comparisons while below are those of the CLM.
The numbers next to each box plot indicates which submodels the comparison is conducted between, i.e. $1,2$ means the results is for the gradient similarities range between submodels 1 and 2. 
The blue dashed vertical line is the average of all ULM submodels similarities and the brown dashed vertical line is the average of the CLM submodels.
The first thing you can notice is the medians of the similarities in the CLM which got shifted more towards 0.
This means, that the saliency maps similarities are more orthogonal between the submodels in the CLM. 
The CLM similarities outliers are less dense than those of the ULM which indicates more similarities are now found within the interquartile range.
Since the attacks crafted by a submodel selected at random have a high probability of being executed on a different submodel, then it is important that the saliency maps generated for the same input by the defending submodel to be more orthogonally similar.
In this case, the attack would have less probability of success.
If the generated saliency maps by the defending submodel is similar to the one crafted the attack,  then the attack has a higher probability of success.
That's why the attack will almost always succeed if the submodel selected at random is the same one that is used to craft the attack.

From Fig.\ref{fig:box}, consider the similarity between submodels 2 and 5 from the CLM (left of the red line) which has the closest median to 0 and the least amount of outliers.  
We show that by using the combination of these two submodels (2 and 5), one for crafting the attack and the other for classifying it ($2 \rightarrow 5$), we achieve the highest robustness among all the other submodels combinations since their gradient similarity is the most orthogonal.
Submodel 2 performed best among the other submodels in the CLM, whether classifying other attacks crafted by other submodels or crafting an attacks with low probability of success.
The combination of ($1 \rightarrow 9$) has the furthest median from the zero and therefore performed worst with a robustness of $22\%$.
The ULM submodels performed similarly with the same range of robustness.
Notice that their saliency maps similarities are within the same range which explains the comparability in the performances.

\section{Conclusion}
In this work, we introduce a unique approach to countering adversarial attacks. 
A CLM consists of a set of submodels that share the same architecture and trained on the same dataset but diverse in their weights values. 
This enhances the robustness of all submodels within the CLM, reducing the probability of crafted attacks to successfully execute on the defending submodels selected at random.
Our argument is that due to population diversity, an attacker will have a non-trivial task of distinguishing which submodel will be dispatched next for the job.
CLM achieves state-of-the-art robustness against several types of attacks when trained with adversarial samples.
Future work would focus on optimizing the model, analyze the defender and attacker next step from a game theory perspective, or crafting a successful attack for this model.

\bibliographystyle{unsrt}  
\bibliography{clm}

\begin{thebibliography}{10}

\bibitem{szegedy2013intriguing}
Christian Szegedy, Wojciech Zaremba, Ilya Sutskever, Joan Bruna, Dumitru Erhan,
  Ian Goodfellow, and Rob Fergus.
\newblock Intriguing properties of neural networks.
\newblock {\em arXiv preprint arXiv:1312.6199}, 2013.

\bibitem{goodfellow2014explaining}
Ian~J Goodfellow, Jonathon Shlens, and Christian Szegedy.
\newblock Explaining and harnessing adversarial examples.
\newblock {\em arXiv preprint arXiv:1412.6572}, 2014.

\bibitem{madry2017towards}
Aleksander Madry, Aleksandar Makelov, Ludwig Schmidt, Dimitris Tsipras, and
  Adrian Vladu.
\newblock Towards deep learning models resistant to adversarial attacks.
\newblock {\em arXiv preprint arXiv:1706.06083}, 2017.

\bibitem{zhou2018breaking}
Yan Zhou, Murat Kantarcioglu, and Bowei Xi.
\newblock Breaking transferability of adversarial samples with randomness.
\newblock {\em arXiv preprint arXiv:1805.04613}, 2018.

\bibitem{xie2017mitigating}
Cihang Xie, Jianyu Wang, Zhishuai Zhang, Zhou Ren, and Alan Yuille.
\newblock Mitigating adversarial effects through randomization.
\newblock {\em arXiv preprint arXiv:1711.01991}, 2017.

\bibitem{guo2017countering}
Chuan Guo, Mayank Rana, Moustapha Cisse, and Laurens Van Der~Maaten.
\newblock Countering adversarial images using input transformations.
\newblock {\em arXiv preprint arXiv:1711.00117}, 2017.

\bibitem{liu2018towards}
Xuanqing Liu, Minhao Cheng, Huan Zhang, and Cho-Jui Hsieh.
\newblock Towards robust neural networks via random self-ensemble.
\newblock In {\em Proceedings of the European Conference on Computer Vision
  (ECCV)}, pages 369--385, 2018.

\bibitem{lecuyer2019certified}
Mathias Lecuyer, Vaggelis Atlidakis, Roxana Geambasu, Daniel Hsu, and Suman
  Jana.
\newblock Certified robustness to adversarial examples with differential
  privacy.
\newblock In {\em 2019 IEEE Symposium on Security and Privacy (SP)}, pages
  656--672. IEEE, 2019.

\bibitem{li2018certified}
Bai Li, Changyou Chen, Wenlin Wang, and Lawrence Carin.
\newblock Certified adversarial robustness with additive noise.
\newblock {\em arXiv preprint arXiv:1809.03113}, 2018.

\bibitem{dhillon2018stochastic}
Guneet~S Dhillon, Kamyar Azizzadenesheli, Zachary~C Lipton, Jeremy Bernstein,
  Jean Kossaifi, Aran Khanna, and Anima Anandkumar.
\newblock Stochastic activation pruning for robust adversarial defense.
\newblock {\em arXiv preprint arXiv:1803.01442}, 2018.

\bibitem{xu2017feature}
Weilin Xu, David Evans, and Yanjun Qi.
\newblock Feature squeezing: Detecting adversarial examples in deep neural
  networks.
\newblock {\em arXiv preprint arXiv:1704.01155}, 2017.

\bibitem{xu2017feature2}
Weilin Xu, David Evans, and Yanjun Qi.
\newblock Feature squeezing mitigates and detects carlini/wagner adversarial
  examples.
\newblock {\em arXiv preprint arXiv:1705.10686}, 2017.

\bibitem{liao2018defense}
Fangzhou Liao, Ming Liang, Yinpeng Dong, Tianyu Pang, Xiaolin Hu, and Jun Zhu.
\newblock Defense against adversarial attacks using high-level representation
  guided denoiser.
\newblock In {\em Proceedings of the IEEE Conference on Computer Vision and
  Pattern Recognition}, pages 1778--1787, 2018.

\bibitem{athalye2018robustness}
Anish Athalye and Nicholas Carlini.
\newblock On the robustness of the cvpr 2018 white-box adversarial example
  defenses.
\newblock {\em arXiv preprint arXiv:1804.03286}, 2018.

\bibitem{raghunathan2018certified}
Aditi Raghunathan, Jacob Steinhardt, and Percy Liang.
\newblock Certified defenses against adversarial examples.
\newblock {\em arXiv preprint arXiv:1801.09344}, 2018.

\bibitem{raghunathan2018semidefinite}
Aditi Raghunathan, Jacob Steinhardt, and Percy Liang.
\newblock Semidefinite relaxations for certifying robustness to adversarial
  examples.
\newblock {\em arXiv preprint arXiv:1811.01057}, 2018.

\bibitem{wong2018provable}
Eric Wong and Zico Kolter.
\newblock Provable defenses against adversarial examples via the convex outer
  adversarial polytope.
\newblock In {\em International Conference on Machine Learning}, pages
  5286--5295. PMLR, 2018.

\bibitem{guo2018sparse}
Yiwen Guo, Chao Zhang, Changshui Zhang, and Yurong Chen.
\newblock Sparse dnns with improved adversarial robustness.
\newblock {\em arXiv preprint arXiv:1810.09619}, 2018.

\bibitem{hein2017formal}
Matthias Hein and Maksym Andriushchenko.
\newblock Formal guarantees on the robustness of a classifier against
  adversarial manipulation.
\newblock {\em arXiv preprint arXiv:1705.08475}, 2017.

\bibitem{weng2018evaluating}
Tsui-Wei Weng, Huan Zhang, Pin-Yu Chen, Jinfeng Yi, Dong Su, Yupeng Gao,
  Cho-Jui Hsieh, and Luca Daniel.
\newblock Evaluating the robustness of neural networks: An extreme value theory
  approach.
\newblock {\em arXiv preprint arXiv:1801.10578}, 2018.

\bibitem{xiao2018training}
Kai~Y Xiao, Vincent Tjeng, Nur~Muhammad Shafiullah, and Aleksander Madry.
\newblock Training for faster adversarial robustness verification via inducing
  relu stability.
\newblock {\em arXiv preprint arXiv:1809.03008}, 2018.

\bibitem{wang2018analyzing}
Yizhen Wang, Somesh Jha, and Kamalika Chaudhuri.
\newblock Analyzing the robustness of nearest neighbors to adversarial
  examples.
\newblock In {\em International Conference on Machine Learning}, pages
  5133--5142. PMLR, 2018.

\bibitem{papernot2018deep}
Nicolas Papernot and Patrick McDaniel.
\newblock Deep k-nearest neighbors: Towards confident, interpretable and robust
  deep learning.
\newblock {\em arXiv preprint arXiv:1803.04765}, 2018.

\bibitem{liu2018adv}
Xuanqing Liu, Yao Li, Chongruo Wu, and Cho-Jui Hsieh.
\newblock Adv-bnn: Improved adversarial defense through robust bayesian neural
  network.
\newblock {\em arXiv preprint arXiv:1810.01279}, 2018.

\bibitem{tramer2018ensemble}
Florian {Tramèr}, Alexey {Kurakin}, Nicolas {Papernot}, Ian~J. {Goodfellow},
  Dan {Boneh}, and Patrick~D. {McDaniel}.
\newblock Ensemble adversarial training: Attacks and defenses.
\newblock In {\em 6th International Conference on Learning Representations,
  ICLR 2018}, 2018.

\bibitem{strauss2017ensemble}
Thilo {Strauss}, Markus {Hanselmann}, Andrej {Junginger}, and Holger {Ulmer}.
\newblock Ensemble methods as a defense to adversarial perturbations against
  deep neural networks.
\newblock {\em arXiv preprint arXiv:1709.03423}, 2017.

\bibitem{chen2020adversarial}
Tianlong Chen, Sijia Liu, Shiyu Chang, Yu~Cheng, Lisa Amini, and Zhangyang
  Wang.
\newblock Adversarial robustness: From self-supervised pre-training to
  fine-tuning.
\newblock In {\em Proceedings of the IEEE/CVF Conference on Computer Vision and
  Pattern Recognition}, pages 699--708, 2020.

\bibitem{wang2020advms}
Xiao Wang, Siyue Wang, Pin-Yu Chen, Xue Lin, and Peter Chin.
\newblock Advms: A multi-source multi-cost defense against adversarial attacks.
\newblock In {\em ICASSP 2020-2020 IEEE International Conference on Acoustics,
  Speech and Signal Processing (ICASSP)}, pages 2902--2906. IEEE, 2020.

\bibitem{wang2019protecting}
Xiao Wang, Siyue Wang, Pin-Yu Chen, Yanzhi Wang, Brian Kulis, Xue Lin, and
  Peter Chin.
\newblock Protecting neural networks with hierarchical random switching:
  Towards better robustness-accuracy trade-off for stochastic defenses.
\newblock {\em arXiv preprint arXiv:1908.07116}, 2019.

\bibitem{lecun1998gradient}
Yann LeCun, L{\'e}on Bottou, Yoshua Bengio, and Patrick Haffner.
\newblock Gradient-based learning applied to document recognition.
\newblock {\em Proceedings of the IEEE}, 86(11):2278--2324, 1998.

\bibitem{he2016deep}
Kaiming He, Xiangyu Zhang, Shaoqing Ren, and Jian Sun.
\newblock Deep residual learning for image recognition.
\newblock In {\em Proceedings of the IEEE conference on computer vision and
  pattern recognition}, pages 770--778, 2016.

\bibitem{croce2020reliable}
Francesco Croce and Matthias Hein.
\newblock Reliable evaluation of adversarial robustness with an ensemble of
  diverse parameter-free attacks.
\newblock In {\em International conference on machine learning}, pages
  2206--2216. PMLR, 2020.

\bibitem{dong2018boosting}
Yinpeng Dong, Fangzhou Liao, Tianyu Pang, Hang Su, Jun Zhu, Xiaolin Hu, and
  Jianguo Li.
\newblock Boosting adversarial attacks with momentum.
\newblock In {\em Proceedings of the IEEE conference on computer vision and
  pattern recognition}, pages 9185--9193, 2018.

\bibitem{athalye2018synthesizing}
Anish Athalye, Logan Engstrom, Andrew Ilyas, and Kevin Kwok.
\newblock Synthesizing robust adversarial examples.
\newblock In {\em International conference on machine learning}, pages
  284--293. PMLR, 2018.

\bibitem{kim2020torchattacks}
Hoki Kim.
\newblock Torchattacks: A pytorch repository for adversarial attacks.
\newblock {\em arXiv preprint arXiv:2010.01950}, 2020.

\bibitem{pang2018towards}
Tianyu Pang, Chao Du, Yinpeng Dong, and Jun Zhu.
\newblock Towards robust detection of adversarial examples.
\newblock {\em Advances in Neural Information Processing Systems}, 31, 2018.

\bibitem{pang2019improving}
Tianyu Pang, Kun Xu, Chao Du, Ning Chen, and Jun Zhu.
\newblock Improving adversarial robustness via promoting ensemble diversity.
\newblock In {\em International Conference on Machine Learning}, pages
  4970--4979. PMLR, 2019.

\bibitem{zhou2021exploring}
Yan Zhou, Murat Kantarcioglu, and Bowei Xi.
\newblock Exploring the effect of randomness on transferability of adversarial
  samples against deep neural networks.
\newblock {\em IEEE Transactions on Dependable and Secure Computing}, 2021.

\bibitem{bogun2021saliency}
Alex Bogun, Dimche Kostadinov, and Damian Borth.
\newblock Saliency diversified deep ensemble for robustness to adversaries.
\newblock {\em arXiv preprint arXiv:2112.03615}, 2021.

\end{thebibliography}

\end{document}